\documentclass{article}

\usepackage[round]{natbib}
\usepackage[colorlinks=true,allcolors=blue]{hyperref}
\usepackage{microtype}
\usepackage{graphicx}
\usepackage{subfig}
\usepackage{booktabs}
\usepackage{amsmath,amssymb,graphicx}
\usepackage{arxiv}

\title{A Fully Differentiable Beam Search Decoder}

\def\X{{\mathbf X}}

\newcommand{\R}{\mathbb{R}}
\DeclareMathOperator*{\logadd}{logadd}
\DeclareMathOperator*{\argmax}{argmax}
\newcommand{\voireq}[1]{Equation~(\ref{#1})}
\newcommand{\voirsec}[1]{Section~\ref{#1}}
\newcommand{\voirfig}[1]{Figure~\ref{#1}}
\newcommand{\voirtbl}[1]{Table~\ref{#1}}

\author{
  Ronan Collobert\\
  Facebook AI Research
  \And
  Awni Hannun\\
  Facebook AI Research
  \And
  Gabriel Synnaeve\\
  Facebook AI Research
}

\begin{document}

\maketitle

\begin{abstract}
  We introduce a new beam search decoder that is fully differentiable, making
  it possible to optimize at training time through the inference procedure.
  Our decoder allows us to combine models which operate at different
  granularities (e.g. acoustic and language models). It can be used when
  target sequences are not aligned to input sequences by considering all
  possible alignments between the two. We demonstrate our approach scales by
  applying it to speech recognition, jointly training acoustic and word-level
  language models. The system is end-to-end, with gradients flowing through
  the whole architecture from the word-level transcriptions. Recent research
  efforts have shown that deep neural networks with attention-based
  mechanisms are powerful enough to successfully train an acoustic model from
  the final transcription, while implicitly learning a language model.
  Instead, we show that it is possible to discriminatively train an acoustic
  model jointly with an \emph{explicit} and possibly pre-trained language
  model.
\end{abstract}

\section{Introduction}

End-to-end models for tasks such as automatic speech recognition require the
use of either structured loss functions like Connectionist Temporal
Classification~\citep{graves2006} or unstructured models like
sequence-to-sequence~\citep{sutskever2014sequence} which leverage an attention
mechanism~\citep{bahdanau2014neural} to learn an implicit alignment. Both these
types of models suffer from an \emph{exposure bias} and a \emph{label bias}
problem.

Exposure-bias results from the mismatch between how these models are trained
and how they are used at inference~\citep{ranzato2016sequence,
wiseman2016sequence, baskar2018promising}. While training, the model is never
exposed to its own mistakes since it uses the ground-truth target as guidance.
At inference the target is unavailable and a beam search is typically used,
often with the constraints of either a lexicon, a transition model, or both.
The model must rely on past predictions to inform future decisions and may not
be robust to even minor mistakes.

Label-bias occurs when locally normalized transition scores are
used~\citep{lecun1998gtn, Lafferty2001, bottou2005graph}. Local normalization causes outgoing
transitions at a given time-step to compete with one another. A symptom of this
is that it can be difficult for the model to revise its past predictions based
on future observations. As a consequence, globally normalized models are
strictly more powerful than locally normalized models~\citep{P16-1231}.

In this work we build a fully differentiable beam search decoder (DBD) which
can be used to efficiently approximate a powerful discriminative model without
either exposure bias or label bias. We alleviate exposure bias by performing a
beam search which includes the constraints from a lexicon and transition
models. In fact, DBD can also jointly train multiple models operating at
different granularities (e.g. token and word-level language models). The DBD
avoids the label bias problem since it uses unnormalized scores and combines
these scores into a global (sequence-level) normalization term.

Our differentiable decoder can handle \emph{unaligned} input and output
sequences where multiple alignments exist for a given output. The DBD model
also seamlessly and efficiently trains token-level, word-level or other
granularity transition models on their own or in conjunction. Connectionist
Temporal Classification (CTC)~\citep{graves2006} can handle unaligned sequences
and is commonly used in automatic speech recognition (ASR)~\citep{amodei2016}
and other sequence labeling tasks~\citep{liwicki2007, huang2016}. The Auto
Segmentation criterion (ASG)~\citep{collobert2016} can also deal with unaligned
sequences. However, neither of these criteria allow for joint training of
arbitrary transition models.

Because DBD learns how to aggregate scores from various models at training
time, it avoids the need for a grid search over decoding parameters (e.g.
language model weights and word insertion terms) on a held-out validation set
at test time. Furthermore, all of the scores used by the differentiable decoder
are \emph{unnormalized} and we thus discard the need for costly normalization
terms over large vocabulary sizes. In fact, when using DBD, training a language
model with a two-million word vocabulary instead of a two-thousand word
vocabulary would incur little additional cost.

We apply DBD to the task of automatic speech recognition and show
competitive performance on the Wall Street Journal (WSJ)
corpus~\citep{douglas1992wsj}. Compared to other baselines which only use
the acoustic data and transcriptions, our model achieves word error rates
which are comparable to state-of-the-art.  We also show that DBD enables
much smaller beam sizes and smaller and simpler models while achieving
lower error rates. This is crucial, for example, in deploying models with
tight latency and throughput constraints.

In the following section we give a description of the exact discriminative
model we wish to learn and in Sections~\ref{sec:decoding} and~\ref{sec:diffdec}
show how a differentiable beam search can be used to efficiently approximate
this model. In \voirsec{sec:diffdec} we also explain the target
sequence-tracking technique which is critical to the success of DBD. We
explain how DBD can be applied to the task of ASR in \voirsec{sec:asr} along
with a description of the acoustic and language models (LMs) we consider.
\voirsec{sec:experiments} describes our experimental setup and results on the
WSJ speech recognition task. In \voirsec{sec:related} we put DBD in context
with prior work and conclude in \voirsec{sec:conclusion}.

\section{Model}
\label{sec:model}

In the following, we consider an input sequence $\X=[X_1,\, \dots,\, X_T]$, and
its corresponding target sequence $\tau$. We also denote a token-level
alignment over the $T$ acoustic frames as $\pi = [\pi_1,\,\dots,\,\pi_T]$. An
alignment leading to the \textit{target} sequence of words $\tau$ is denoted as
$\pi^\tau$. The conditional likelihood of $\tau$ given the input $\X$ is
then obtained by marginalizing over all possible alignments $\pi$ leading to
$\tau$:
\begin{equation}
  \label{eq:marginalization}
  \log P(\tau|\X) = \log \sum_{\forall \pi^\tau} P(\pi^\tau | \X) \,.
\end{equation}

In the following, we consider a \emph{scoring} function $f^t(i|\X)$ which
outputs scores for each frame $t\in [1, \dots, T]$ and each label $i\in {\cal
D}$ in a token set. We also consider a token \emph{transition} model $g(i,j)$
(we stick to a bigram for the sake of simplicity) and a word \emph{transition}
model $h(\cdot)$.  Given an input sequence $\X$, an alignment $\pi^\tau$ is
assigned an unnormalized score $s(\pi^\tau|\X)$ obtained by summing up frame
scores, token-level transition scores along $\pi^\tau$ and the word-level
transition model score:
\begin{equation}
  \label{eq:score}
  s(\pi^\tau|\X) = \sum_{t=1}^T f^t(\pi^\tau_t | \X) + g(\pi^\tau_t|\pi^\tau_{t-1}) + h(\tau)\,.
\end{equation}
It is important to note that the frame scores and transition scores are
all unnormalized. Hence, we do not require any transition model weighting as
the model will learn the appropriate scaling. Also, $P(\pi^\tau|\X)$ in
\voireq{eq:marginalization} is obtained by performing a sequence-level
normalization, applying a softmax over all alignment scores $s(\pi^{\eta}|\X)$
for all possible \emph{valid} sequences of words $\eta$ ($\eta$ necessarily
include $\tau$):
\begin{equation}
  \label{eq:softmax}
  \log P(\pi^\tau|\X) = s(\pi^\tau|\X) - \log \sum_{\forall \eta,\,\forall \pi^{\eta}} e^{s(\pi^{\eta}|\X)}\,.
\end{equation}
Combining \voireq{eq:marginalization} and \voireq{eq:softmax}, and introducing
the operator $\logadd(a,\,b) = \log(e^a+e^b)$ for convenience, our model can be
summarized as:
\begin{equation}
  \label{eq:diffdec}
  \log P(\tau|\X)  =  \logadd_{\forall \pi^\tau} s(\pi^\tau|\X)-\logadd_{\forall \eta,\,\forall \pi^{\eta}} s(\pi^{\eta}|\X)\,.
\end{equation}
Our goal is to optimize jointly the scoring function and the transition models
(token-level and word-level) by maximizing this conditional likelihood over all
labeled pairs $(\X,\, \tau)$ available at training time. In
\voireq{eq:diffdec}, it is unfortunately intractable to exactly compute the
$\logadd()$ over all possible sequences $\eta$ of valid words. In the next
section, we will relate this likelihood to what is computed during decoding at
inference and then show how it can be approximated efficiently. In
\voirsec{sec:diffdec}, we will show how it can be efficiently optimized.

\section{Decoding}
\label{sec:decoding}

At inference, given an input sequence $\X$, one needs to find the best
corresponding word sequence $\tau^\star$. A popular decoding approach is to
define the problem formally as finding $\argmax_{\tau} s(\pi^\tau|\X)$,
implemented as a Viterbi search. However, this approach takes in account
only the best alignment $\pi^\tau$ leading to $\tau$. Keeping in mind the
normalization in~\voireq{eq:diffdec}, and following the footsteps of
\citep{bottou1991phd}, we are interested instead in finding the $\tau$ which
maximizes the Forward score:
\begin{equation}
  \label{eq:decoding}
  \begin{split}
    \max_\tau \log P(\tau|\X) & = \max_\tau \logadd_{\forall \pi^\tau} s(\pi^\tau|\X)\\\
    & \approx \max_{\tau\in {\cal B}} \logadd_{\forall \pi^\tau \in {\cal B}} s(\pi^\tau|\X)\,.
    \end{split}
\end{equation}
The first derivation in \voireq{eq:decoding} is obtained by plugging in
\voireq{eq:diffdec} and noticing that the normalization term
$\logadd_{\forall\eta}()$ is constant with respect to $\tau$. As the search
over all possible sequences of words is intractable, one performs a beam
search, which results in a final set of hypotheses ${\cal B}$. For each
hypothesis in the beam ($\tau\in {\cal B}$), note that only the most
promising alignments leading to this hypothesis will also be in the beam
($\forall \pi^\tau \in {\cal B}$); in contrast to pure Viterbi decoding,
these alignments are aggregated through a $\logadd()$ operation instead of
a $\max()$ operation.

Our beam search decoder uses a word lexicon (implemented as a trie converting
sequences of tokens into words) to constrain the search to only \emph{valid}
sequences of words $\tau$. The decoder tracks hypotheses $\tau$ with the
highest scores by bookkeeping tuples of ``(lexicon state, transition model
state, score)'' as it iterates through time. At each time step, hypotheses with
the same transition model state and lexicon state are merged into the top
scoring hypothesis with this state. The score of the resulting hypothesis is
the $\logadd()$ of the combined hypotheses.

\subsection{Decoding to compute the likelihood normalization}

The normalization term computed over all possible valid sequence of words
$\eta$ in the conditional likelihood \voireq{eq:diffdec} can be efficiently
approximated by the decoder, subject to a minor modification.
\begin{equation}
  \label{eq:decoding-Z}
  \begin{split}
    \logadd_{\forall \eta,\,\forall \pi^{\eta}}s(\pi^{\eta}|\X) & = \logadd_{\forall \eta}\logadd_{\forall \pi^{\eta}}s(\pi^{\eta}|\X) \\
    & \approx \logadd_{\forall \eta \in {\cal B}}\logadd_{\forall \pi^{\eta} \in {\cal B}}s(\pi^{\eta}|\X)\,,
  \end{split}
\end{equation}
where ${\cal B}$ is the set of hypotheses retained by the decoder beam.
Compared to \voireq{eq:decoding}, the only change in \voireq{eq:decoding-Z} is
the \emph{final} ``aggregation'' of the hypotheses in the beam: at inference,
one performs a $\max()$ operation, while to compute the likelihood
normalization one performs a $\logadd()$.

\begin{figure}
  \centering
  \includegraphics[width=0.5\linewidth]{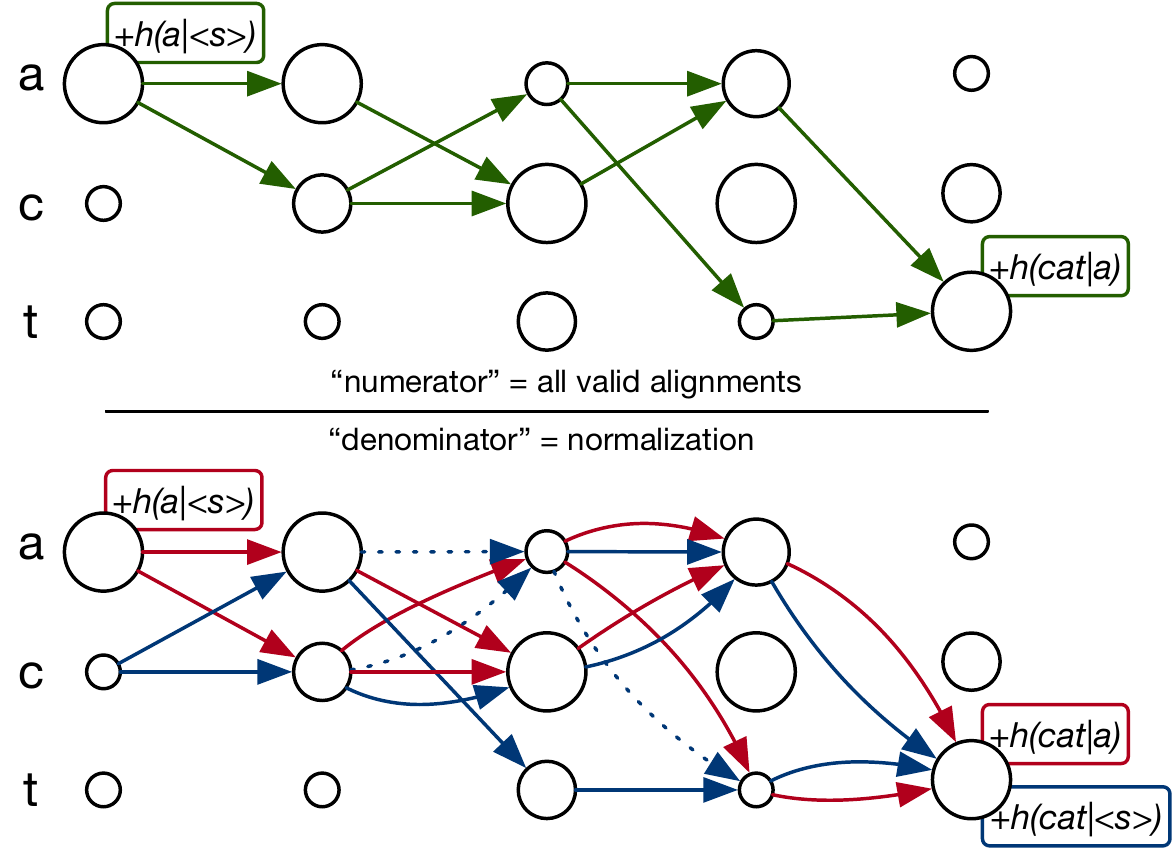}
  \label{fig-dbd-denominator}
  \caption{
    An example of the DBD computation of the loss (\voireq{eq:diffdecapprox}),
    with a target transcription of ``a cat'', using a lexicon \{a, cat\}, 5
    frames in total, and a word-level bigram LM $h(\cdot)$. Circle sizes are
    proportional to the AM score and paths through the graph are aggregated
    with a $\logadd$. The first term (maximized, ``numerator'') corresponds to
    all valid alignments (in green). The second term corresponds to the beam
    search which is used to construct the denominator
    (\voireq{eq:diffdecforreal}). Dashed arrows denote transitions not included
    in the beam. Multiple alignments leading to the same word are merged, and
    the LM scores are added (rounded rectangles) as words are considered in the
    beam.}
\end{figure}

\section{Differentiable Decoding}
\label{sec:diffdec}

As mentioned in \voirsec{sec:model}, we aim to jointly train the scoring
function and the transition models (token-level and word-level) by maximizing
the conditional likelihood in \voireq{eq:diffdec} over all training pairs
$(\X,\, \tau)$. Leveraging the decoder to approximate the normalization term
(with \voireq{eq:decoding-Z}), and denoting $\pi \in {\cal T}_{\tau}$ and
$\pi \in {\cal B}$ for all alignments leading to the target and all alignments
in the beam respectively, this corresponds to maximizing:
\begin{equation}
  \label{eq:diffdecapprox}
  \log P(\tau|\X)  =  \logadd_{\forall \pi \in {\cal T}_{\tau} } s(\pi|\X)-\logadd_{\forall \pi \in {\cal B}} s(\pi|\X)\,.
\end{equation}
This approximated conditional likelihood can be computed and differentiated in
a tractable manner. The $\logadd_{\pi \in {\cal T}_{\tau}}$ can be exactly
evaluated via dynamic programming (the Forward algorithm). The normalization
term $\logadd_{\pi \in {\cal B}}$ is computed efficiently with the decoder.
Interestingly, both of these procedures (the Forward algorithm and decoding)
only invoke additive, multiplicative and $\logadd()$ operations in an iterative
manner. In that respect, everything is fully differentiable, and one can
compute the first derivatives $\partial\log P(\tau|\X)/\partial f_t(i|\X)$,
$\partial\log P(\tau|\X)/\partial g(i|j)$, $\partial\log
P(\tau|\X)/\partial h$ by backpropagating through the Forward and decoder
iterative processes.

\subsection{On the normalization approximation}

In practice, the normalization approximation shown in \voireq{eq:diffdecapprox}
is unfortunately inadequate. Indeed, the beam ${\cal B}$ may or may not
contain some (or all) character alignments found in ${\cal T}_{\tau}$. We
observe that models with improper normalization fail to converge as everything
(scoring function and transition models) outputs unnormalized scores. We correct
the normalization by considering instead the $\logadd()$ over $({\cal
B}\setminus {\cal B}\cap {\cal T}_\tau) \cup {\cal T}_\tau$. This quantity
can be efficiently computed using the following observation:
\begin{equation}
  \label{eq:diffdecforreal}
  \begin{split}
  \logadd_{\forall \pi \in ({\cal B}\setminus {\cal B}\cap {\cal T}_\tau)\cup{\cal T}_\tau } s(\pi|\X) =
  \log(
  & e^{ \logadd_{\forall \pi \in {\cal B}} s(\pi|\X) } \\
  - & e^{ \logadd_{\forall \pi \in {\cal B}\cap{\cal T}_\tau} s(\pi|\X) } \\
  + & e^{ \logadd_{\forall \pi \in {\cal T}_\tau} s(\pi|\X) } )\,.
  \end{split}
\end{equation}
The $\logadd()$ terms over ${\cal B}$ and ${\cal T}_\tau$ are given by the
decoder and the Forward algorithm respectively. The $\logadd()$ term over
${\cal B}\cap{\cal T}_\tau$ can also be computed by the decoder by tracking
alignments in the beam which correspond to the ground truth $\tau$. While
this adds extra complexity to the decoder, it is an essential feature for
successful training.

\subsection{On the implementation}
Our experience shows that implementing an efficient differentiable version of
the decoder is tricky. First, it is easy to miss a term in the gradient given
the complexity of the decoding procedure. It is also difficult to check the
accuracy of the gradients by finite differences (and thus hard to find
mistakes) given the number of operations involved in a typical decoding pass.
To ensure the correctness of our derivation, we first implemented a custom C++
autograd (operating on scalars, as there are no vector operations in the
decoder). We then designed a custom version of the differentiable decoder
(about 10$\times$ faster than the autograd version) which limits memory allocation and
checked the correctness of the gradients via the autograd version.

\section{Application to Speech Recognition}
\label{sec:asr}

In a speech recognition framework, the input sequence $\X$ is an acoustic
utterance, and the target sequence $\tau$ is the corresponding word
transcription. Working at the word level is challenging, as corpora are usually
not large enough to model rare words properly. Also, some words in the
validation or test sets may not be present at training time. Instead of
modeling words, one considers sub-word units -- like phonemes,
context-dependent phonemes, or characters. In this work, we use characters for
simplicity. Given an acoustic sequence $\X$, character-level alignments
corresponding to the word transcription $\tau$ are $\pi^\tau$. The correct
character-level alignment is unknown. The scoring function $f^t(\pi^\tau_t |
\X)$ is an acoustic model predicting character scores at each frame $X_t$ of
the utterance $X$. The transition model $g(i|j)$ learns a character-level
language model, and the word transition model $h(\tau)$ is a word language
model.

Both the acoustic and language models can be customized in our approach. We use
a simple ConvNet architecture (leading to a reasonable end-to-end word error
rate performance) for the acoustic model, and experimented with a few different
language models. We now describe these models in more details.

\subsection{Acoustic Model}
We consider a 1D ConvNet for the acoustic model (AM), with Gated Linear Units
(GLUs)~\citep{dauphin2017} for the transfer function and dropout as
regularization. Given an input $\X^l$, the $l^{th}$ layer of a Gated ConvNet computes
$$
h_l(\X^l) = (\X^l\ast \mathbf{W}^l + \mathbf{b})\otimes \sigma(\X^l\ast \mathbf{W'}^l+\mathbf{b'})\,,
$$
where $\mathbf{W}^l$, $\mathbf{b}$ and $\mathbf{W'}^l$, $\mathbf{b'}$
are trainable parameters of two different convolutions. As $\sigma()$ is
the sigmoid function, and $\otimes$ is the element-wise product between
matrices, GLUs can be viewed as a gating mechanism to train deeper
networks.
Gated ConvNets have been shown to perform well on a number of
tasks, including speech recognition~\citep{liptchinsky2017}. As the
differentiable decoder requires heavy compute, we bootstrapped the training of
the acoustic model with ASG. The ASG criterion is similar to
\voireq{eq:softmax} but the normalization term is taken over \emph{all}
sequences of tokens
\begin{equation}
  \label{eq:asg}
  \log P(\pi^\tau|\X) = s(\pi^\tau|\X) - \log \sum_{\forall \pi} e^{s(\pi|\X)}
\end{equation}
with the alignment score is given by
\begin{equation}
  \label{eq:asg_score}
  s(\pi^\tau|\X) = \sum_{t=1}^T f^t(\pi^\tau_t | \X) + g(\pi^\tau_t|\pi^\tau_{t-1})
\end{equation}
which does not include a word language model.

\subsection{Language Models}
\label{sec:lms}

The character language model as shown in \voireq{eq:score} was chosen to be
a simple trainable scalar $g(i|j)=\lambda_{ij}$. We experimented with
several word language models:
\begin{enumerate}
\item A \emph{zero} language model $h(\tau)=0$. This special case is a
  way to evaluate how knowing the lexicon can help the acoustic model
  training. Indeed, even when there is no language model information, the
  normalization shown in \voireq{eq:diffdec} still takes in account the
  available lexicon. Only sequences of letters $\pi$ leading to a valid
  sequence of words $\tau$ are considered (compared to any sequence of
  letters, as in ASG or LF-MMI).
\item A \emph{pre-trained} language model, possibly on data not available
  for the acoustic model training. We considered in this case
  \begin{equation}
    \label{eq:lmpretrained}
    h(\tau) = \lambda \log P_{lm}(\tau) + \gamma\,,
  \end{equation}
  where $P_{lm}(\tau)$ is the pre-trained language model. The language
  model weight $\lambda$ and word insertion score $\gamma$ are parameters
  \emph{trained} jointly with the acoustic model.
\item A \emph{bilinear} language model. Denoting the sequence of words
  $\tau=[\tau_1,\,\dots,\,\tau_N]$, we consider the unnormalized
  language model score:
  \begin{equation}
    h(\tau) = w_{\tau_N}^{T} \sum_{n=1}^{K} \mathbf{M}_n w_{\tau_{N-n}}\,,
  \end{equation}
  where $K\geq2$ is the order of the language model. The word embeddings
  $w_i\in\R^d$ ($d$ to be chosen) and the projection matrices $M_i\in
  \R^{d\times d}$ are trained jointly with the acoustic model. It is
  worth mentioning that the absence of normalization makes this
  particular language model efficient.
\end{enumerate}

\section{Experiments}
\label{sec:experiments}

\begin{figure*}
  \centering
  \subfloat[]{\label{fig:baseline-cer}\includegraphics[width=0.24\linewidth]{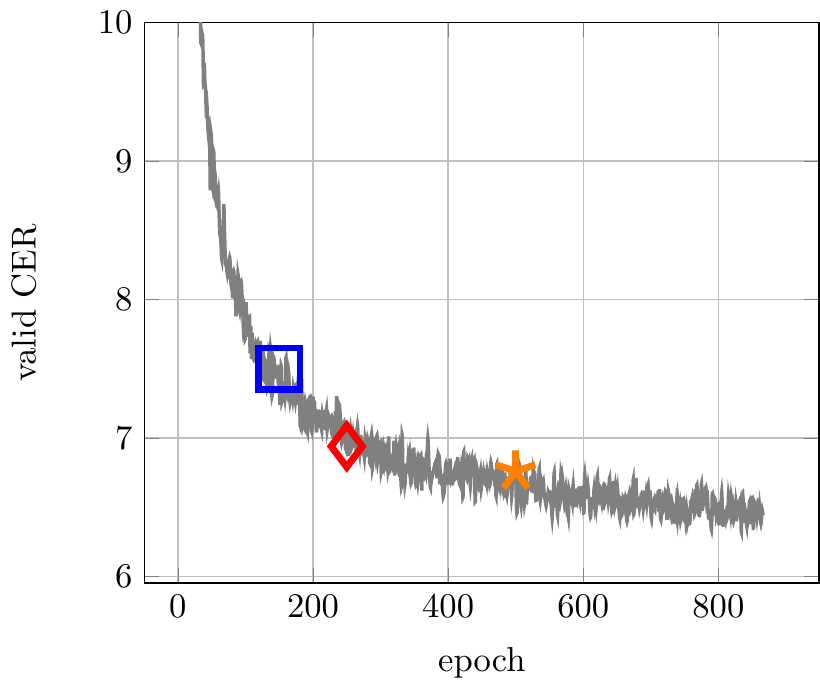}}
  \hspace*{0.1cm}
  \subfloat[]{\label{fig:baseline-wer}\includegraphics[width=0.24\linewidth]{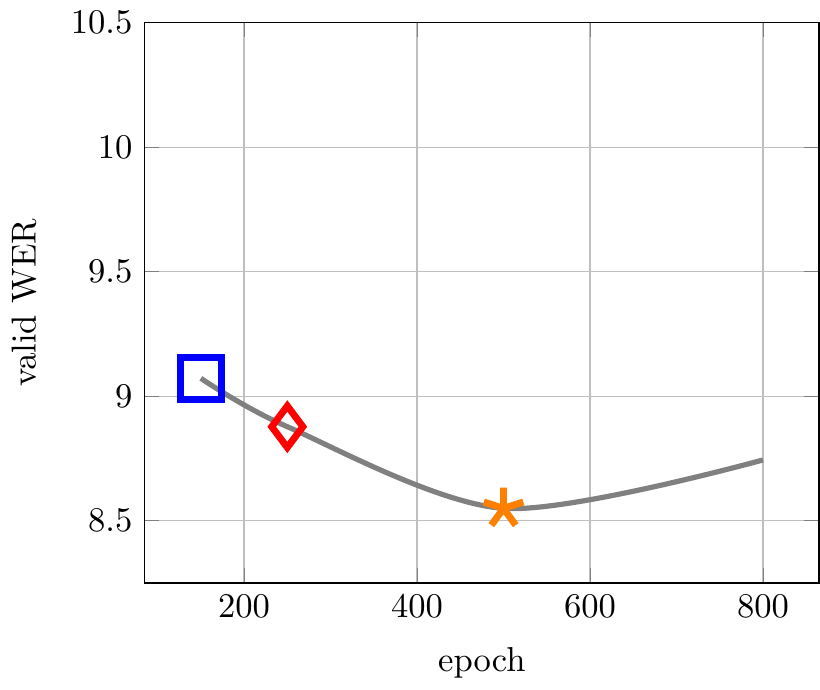}}
  \hspace*{0.1cm}
  \subfloat[]{\label{fig:dbd-convergence-train}\includegraphics[width=0.24\linewidth]{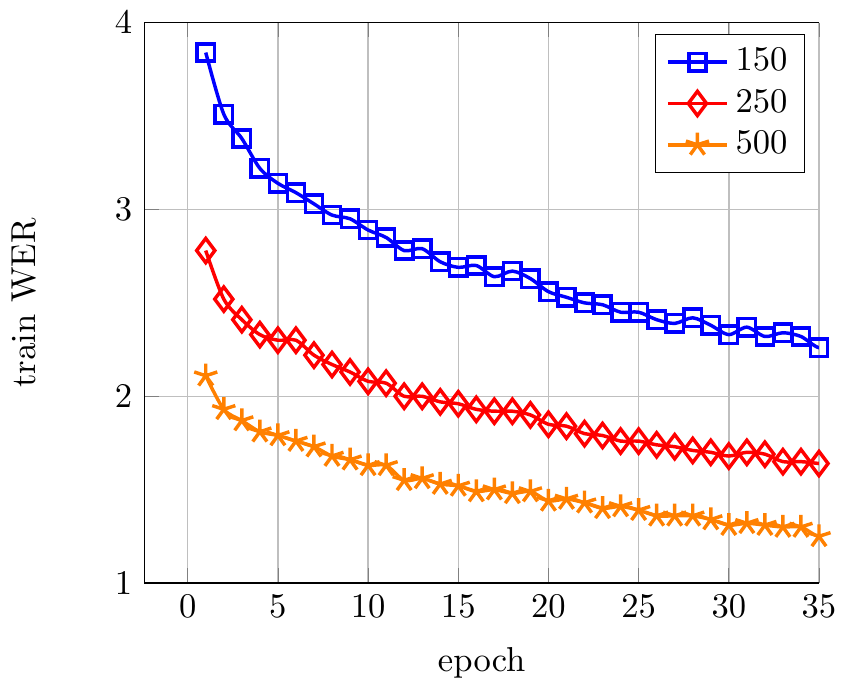}}
  \hspace*{0.1cm}
  \subfloat[]{\label{fig:dbd-convergence-valid}\includegraphics[width=0.24\linewidth]{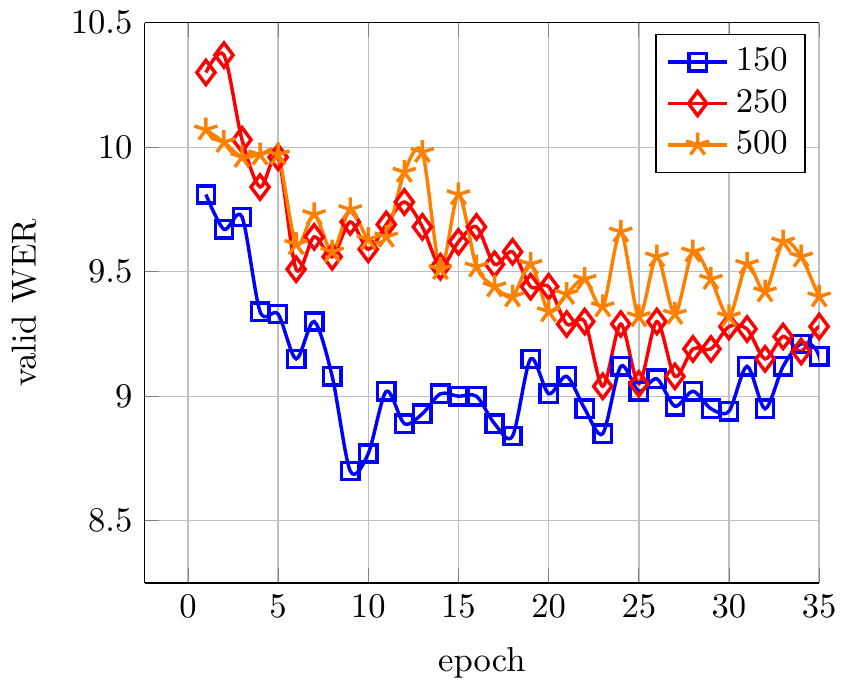}}
  \caption{
    \label{fig-baseline-cer-wer}
    Performance of the 10M parameters ConvNet, with respect to training epochs
    over WSJ. ASG baseline valid error is shown CER (a) and WER (b). Each ASG
    WER was obtained by running a full grid search (beam size 8000). DBD train
    (c) and valid (d) WER: three DBD runs were forked from the ASG baseline at
\textit{epochs} 150, 250 and 500 (see colors/shapes) and trained with beam
size 500.}
\end{figure*}

We performed experiments with WSJ (about $81$h of transcribed audio data). We
consider the standard subsets \emph{si284}, \emph{nov93dev} and
\emph{nov92} for training, validation and test, respectively. We use
log-mel filterbanks as features fed to the acoustic model, with 40 filters
of size 25ms, strided by 10ms. We consider an end-to-end setup, where the
token set ${\cal D}$ (see \voirsec{sec:model}) includes English
letters (a-z), the apostrophe and the period character, as well as a
space character, leading to 29 different tokens. No data augmentation
or speaker adaptation was performed. As WSJ ships with both acoustic
training data and language-model specific training data, we consider two
different training setups:
\begin{enumerate}
\item Language models are pre-trained (see \voireq{eq:lmpretrained}) with
  the full available language model data. This allows us to
  demonstrate that our approach can tune automatically the language model
  weight and leverage the language model information during the training
  of the acoustic model.
\item Both acoustic and language models are trained with acoustic (and
  corresponding transcription) data only. This allows us to compare with
  other end-to-end work where only the acoustic data was used.
\end{enumerate}
Pre-trained language models are n-gram models trained with
KenLM~\citep{heafield2011}. The word dictionary contains words from both
the acoustic and language model data. We did not perform any thresholding,
leading to about 160K distinct words.

All the models are trained with stochastic gradient descent (SGD), enhanced
with gradient clipping~\citep{pascanu2013rnn} and weight
normalization~\citep{salimans2016wn}. In our experience, these two improvements
over vanilla SGD allow higher learning rates, and lead to faster and more
stable convergence. Without weight normalization we found GLU-ConvNets very
challenging to train. We use batch training (16 utterances at once), sorting
inputs by length prior to batching for efficiency. Both the neural network
acoustic model and the ASG criterion run on a single GPU. The DBD criterion is
CPU-only. With ASG, a single training epoch over WSJ takes just a few minutes,
while it takes about an hour with DBD.

\begin{figure}
  \centering
  \subfloat[]{\label{fig-X1}\includegraphics[width=0.24\linewidth]{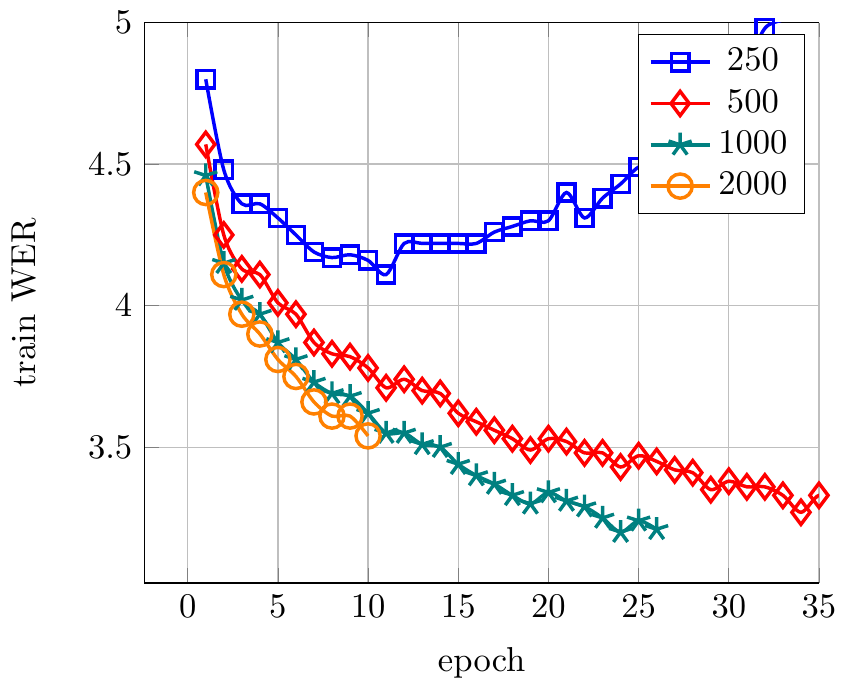}}
  \subfloat[]{\label{fig-X2}\includegraphics[width=0.24\linewidth]{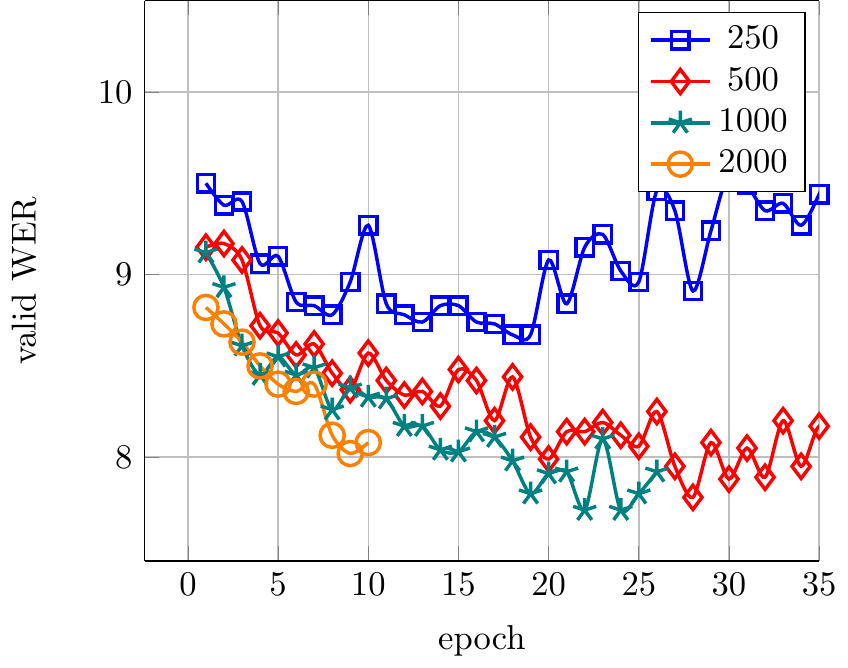}}
  \caption{
    \label{fig:beamsize}
    Training DBD with different \textit{beam size}, showing train (a) and valid
    (b) WER, for the 7.5M parameters model on WSJ.
 }
\end{figure}

\subsection{Leveraging Language-Model Data}

In speech recognition, it is typical to train the acoustic model and the
language model separately. The language model can take advantage of
large text-only corpora. At inference, both models are combined through the
decoding procedure (maximizing \voireq{eq:decoding}). Hyper-parameters
combining the language model (as in \voireq{eq:lmpretrained}) are tuned
through a validation procedure (e.g. grid-search).

We first performed an extensive architecture search, training models with
ASG and selecting a baseline model as the one leading to the best validation
character error rate (CER). Our best ConvNet acoustic model has 10M
parameters and an overall receptive field of 1350ms. A 4-gram language
model was trained over the text data provided with WSJ. A decoding grid search
was then performed at several points during the acoustic model training
\voirfig{fig:baseline-wer}). While CER and WER are correlated, the
best WER did not correspond to the best CER (see \voirfig{fig:baseline-cer}
and \voirfig{fig:baseline-wer}).

\begin{table}
  \caption{\label{tbl:withlmdata} Comparing WER performance of ASG with decoding
    grid-search, and DBD, on WSJ. We compare with standard end-to-end approaches, for reference.
    }
  \begin{center}
    \begin{tabular}{ccccc}
      \toprule
      Model & nov93dev & nov92 \\
      \midrule
      ASG \emph{10M AM} \begin{footnotesize}\textit{(beam size 8000)}\end{footnotesize}  & 8.5 & 5.6 \\
      ASG \emph{10M AM} \begin{footnotesize}\textit{(beam size 500)}\end{footnotesize}  & 8.9 & 5.7 \\
      ASG \emph{7.5M AM} \begin{footnotesize}\textit{(beam size 8000)}\end{footnotesize} & 8.8 & 6.0 \\
      ASG \emph{7.5M AM} \begin{footnotesize}\textit{(beam size 500)}\end{footnotesize} & 9.4 & 6.1 \\
      \cmidrule(lr){1-3}
      DBD \emph{10M AM} \begin{footnotesize}\textit{(beam size 500)}\end{footnotesize} & 8.7 & 5.9 \\
      DBD \emph{7.5M AM} \begin{footnotesize}\textit{(beam size 500)}\end{footnotesize} & 7.7 & 5.3 \\
      DBD \emph{7.5M AM} \begin{footnotesize}\textit{(beam size 1000)}\end{footnotesize} & 7.7 & 5.1 \\
      \cmidrule(lr){1-3}
      Attention RNN+CTC & & 9.3 \\
      \begin{footnotesize}\textit{(3gram)}~\citep{bahdanau2016icassp}\end{footnotesize} & & \\
      CNN+ASG & 9.5 & 5.6 \\
      \begin{footnotesize}\textit{(4-gram)}~\citep{zeghidour2018fully}\end{footnotesize} & & \\
      CNN+ASG \begin{footnotesize}\textit{(wav+convLM)}\end{footnotesize} & 6.8 & 3.5 \\
      \begin{footnotesize}\citep{zeghidour2018fully}\end{footnotesize} & & \\
      RNN+E2E-LF-MMI \begin{footnotesize}\textit{(data augm.)}\end{footnotesize} & & 4.1 \\
      \begin{footnotesize}\textit{(+RNN-LM)}~\citep{hadian2018}\end{footnotesize} & & \\
      BLSTM+PAPB+CE & & 3.8 \\
      \begin{footnotesize}\textit{(RNN-LM)}~\citep{baskar2018promising}\end{footnotesize} & & \\
      \bottomrule
    \end{tabular}
  \end{center}
\end{table}

As DBD training is time-consuming compared to ASG-training, we bootstrapped
several DBD models from three different checkpoints of our ASG model (at epoch
125, 250 and 500). With DBD, the acoustic model is jointly fine-tuned with the
weights of the language model shown in \voireq{eq:lmpretrained}.
\voirfig{fig:dbd-convergence-train} and \voirfig{fig:dbd-convergence-valid}
show the training and validation WER with respect to number of epochs over WSJ.
DBD converges quickly from the pre-trained ASG model, while many epochs (and a
grid-search for the language model hyper-parameters) are required to match the
same WER with ASG. When starting from later ASG epochs (250 and 500), DBD badly
overfits to the training set.

To mitigate overfitting, we trained a variant of our 10M model where the
receptive field was reduced to 870ms (leading to 7.5M
parameters). \voirtbl{tbl:withlmdata} summarizes our results. While ASG is
unable to match the WER performance of the 1450ms receptive field model,
training with DBD leads to better performance, demonstrating the advantage
of jointly training the acoustic model with the language model. Not only
does DBD allow for more compact acoustic models, but also DBD-trained
models require much smaller beam at decoding, which brings a clear speed
advantage at inference.

\voirfig{fig:beamsize} shows the effect of the beam size during DBD
training.  Having too small a beam leads to model divergence. Training
epoch times for beam sizes of 500, 1000 and 2000 are 35min, 90m and 180min
respectively. In most experiments, we use a beam size of 500, as larger
beam sizes led to marginal WER improvements. Although, in contrast to
pipelines suffering from exposure bias \citep[e.g.][]{W17-3204}, larger
beams are always better.

\subsection{Experimenting with Acoustic-Only Data}

Recent work on WSJ has shown that end-to-end approaches are good at modeling
acoustic data. Some of these works also demonstrated that with architectures
powerful enough to capture long-range dependencies, end-to-end approaches can
also implicitly model language, and push the WER even further down. DBD allows
us to explicitly design acoustic and language models while training them
jointly. We show in this section that with simple acoustic and language models,
we can achieve WERs on par with existing approaches trained on acoustic-only
data.

\begin{table}
  \caption{\label{tab:exptrain} WSJ performance (WER), using only the
    acoustic training data. ASG n-gram decoding hyper-parameters were tuned
    via grid-search. Beam size for both ASG and DBD was 500. Larger
    beam sizes with ASG did not lead to significant improvements.
  }
  \begin{center}
    \begin{tabular}{ccc}
      \toprule
      Model & nov93dev & nov92 \\
      \midrule
      ASG \emph{(zero LM decoding)} & 18.3 & 13.2 \\
      ASG \emph{(2-gram LM decoding)} & 14.8 & 11.0\\
      ASG \emph{(4-gram LM decoding)} & 14.7 & 11.3\\
      \cmidrule(lr){1-3}
      DBD \emph{zero LM} & 16.9 & 11.6 \\
      DBD \emph{2-gram LM} & 14.6 & 10.4 \\
      DBD \emph{2-gram-bilinear LM} & 14.2 & 10.0 \\
      DBD \emph{4-gram LM} & 13.9 & 9.9 \\
      DBD \emph{4-gram-bilinear LM} & 14.0 & 9.8 \\
      \cmidrule(lr){1-3}
      RNN+CTC & & 30.1 \\
      \begin{footnotesize}\textit{}~\citep{graves2014towards}\end{footnotesize} & & \\
      Attention RNN+CTC & & 18.6 \\
      \begin{footnotesize}\textit{}~\citep{bahdanau2016icassp}\end{footnotesize} & & \\
      Attention RNN+CTC+TLE & & 17.6 \\
      \begin{footnotesize}\textit{}~\citep{bahdanau2016iclr}\end{footnotesize} & & \\
      Attn. RNN+seq2seq+CNN & & 9.6 \\
      \begin{footnotesize}\textit{(speaker adapt.)}~\citep{chan2016latent}\end{footnotesize} & & \\
      BLSTM+PAPB+CE & & 10.8 \\
      \begin{footnotesize}\textit{}~\citep{baskar2018promising}\end{footnotesize} & & \\
      \bottomrule
    \end{tabular}
  \end{center}
\end{table}

In Table~\ref{tab:exptrain} we report standard baselines for this setup, as
well as our own ASG baseline model, decoded with an n-gram trained only on
acoustic data. We compare with DBD-trained models using the three different
language models introduced in \voirsec{sec:lms}: (i)~a zero language model,
which allows us to leverage the word lexicon while training; (ii)~n-gram
language models, pre-trained on acoustic data (where the weighting is trained
jointly with the acoustic model) and (iii) bilinear language models (where all
the parameters are trained jointly with the acoustic model). Results show that
only knowing the lexicon when training the acoustic model already greatly
improves the WER over the baseline ASG model, where the lexicon is known only
at test time. Jointly training a word language model with the acoustic model
further reduces the WER.

\section{Related Work}
\label{sec:related}

Our differentiable decoder belongs to the class of sequence-level training
criteria, which includes Connectionist Temporal Classification
(CTC)~\citep{graves2006,graves2014towards} and the Auto Segmentation (ASG)
criterion~\citep{collobert2016}, as well as Minimum Bayes Risk (MBR and sMBR)~\citep{goel2000minimum,gibson2006smbr,sak2015fast,prabhavalkar2018minimum} and the Maximum Mutual Information
(MMI) criterion~\citep{bahl1986}, amongst others. MMI and ASG are the
closest to our differentiable decoder as they perform global (sequence-
level) normalization, which should help alleviate the \textit{label bias}
problem \citep{lecun1998gtn,Lafferty2001,bottou2005graph,P16-1231}.

Both MBR and MMI are usually trained after (or mixed with) another sequence
loss or a force-alignment phase. MMI maximizes the average mutual
information between the observation $\X$ and its correct transcription
$\tau$. Considering an HMM with states $S_\tau = [s^\tau_1,\,\dots,\,
  s^\tau_T]$ for a given transcription $\tau$, MMI maximizes:
\begin{equation}
  \label{eq:mmi}
I(\tau,\X) = \log P(\X|S_\tau) P(\tau) - \logadd_{\eta} \log P(\X|S_{\eta}) P(\eta)\,.
\end{equation}
In contrast to MMI, MBR techniques integrate over plausible transcriptions
$\eta$, weighting each candidates by some accuracy $A(\tau,\,\eta)$ to
the ground truth -- computing $\sum_{\eta} e^{I(\eta,\X)} A(\tau,\,
\eta)$.

In a neural network context, one can apply Bayes' rule to plug in the
emission probabilities. Ignoring $P(\X)$, the term $\log P(\X|s_t)$ in
\voireq{eq:mmi} is approximated by $\log P(s_t|\X) - \log P(s)$, where
$\log P(s_t|\X) = f_t(s_t|\X)$ (the emissions being normalized per frame),
and $P(s)$ is estimated with the training data. Apart from this approximation,
two differences with our differentiable decoder are critical:
\begin{itemize}
    \item MMI considers normalized probabilities for both the acoustic and
        language model, while we consider unnormalized scores everywhere.
    \item MMI does not jointly train the acoustic and language models. MMI
        does come in different flavors though, with (fixed) token level (phone)
        language models and no lexicon, as found in lattice-free MMI (LF-MMI)
        \citep{povey2016purely}, and even trained end-to-end without any
        alignments (EE-LF-MMI) as in \citep{hadian2018}, with a pre-trained
        phone LM still.
\end{itemize}

ASG maximizes \voireq{eq:asg} which is similar to \voireq{eq:diffdec} but with
two critical differences: (1)~there is no word language model in
\voireq{eq:score}, and (2)~the normalization term is not constrained to valid
sequences of words but is over all possible sequences of letters, and thus can
be computed exactly (as is the case for LF-MMI). Unlike ASG, CTC assumes output
tokens are conditionally independent given the input and includes an optional
\emph{blank} which makes the graph less regular~\citep{liptchinsky2017}.

Because our work is end-to-end, it is also related to seq2seq learning
\citep{sutskever2014sequence,bahdanau2014neural,chan2016latent,wiseman2016sequence,gehring17a},
and in particular training with existing/external language models
\citep{sriram2018cold}. Closest to our work is \citep{baskar2018promising}
that shares a similar motivation, training an acoustic model through beam
search although its (1) loss includes an error rate (as MBR), (2) they
consider partial hypotheses (promising accurate prefix boosting: PAPB),
and in practice (3) they optimize a loss composing this beam search
sequence score with the cross-entropy (CE).

In NLP, training with a beam search procedure is not new
\citep{P04-1015,daume2005learning,wiseman2016sequence}.
Of those, \citep{wiseman2016sequence} is the closest to our work,
training a sequence model through a beam-search with a global sequence score.
To our knowledge, we are the first to train through a beam search decoder for
speech recognition, where the multiplicity of gold transcription alignments
makes the search more complex. Also related, several works are targeting the
loss/evaluation mismatch (and sometimes exposure bias) through reinforcement
learning (policy gradient) \citep{bahdanau2016iclr,ranzato2016sequence} even in
speech recognition \citep{zhou2018improving}.

Finally, our work makes a generic beam search differentiable end-to-end and so
relates to relaxing the beam search algorithm itself (e.g. getting a soft beam
through a soft argmax) \citep{goyal2018continuous}, although we use a discrete
beam. Compared to differentiable dynamic programming
\citep{mensch2018differentiable,bahdanau2016iclr}, we use a $\logadd$ where they
use a softmax and we keep track of an n-best (beam) set while they use
Viterbi-like algorithms.

\section{Conclusion}
\label{sec:conclusion}

We build a fully differentiable beam search decoder which is capable of jointly
training a scoring function and arbitrary transition models. The DBD can handle
unaligned sequences by considering all possible alignments between the input
and the target. Key to this approach is a carefully implemented and highly
optimized beam search procedure which includes a novel target sequence-tracking
mechanism and an efficient gradient computation. As we show, DBD can scale to
very long sequences with thousands of time-steps. We are able to perform a full
training pass (epoch) through the WSJ data in about half-an-hour with a beam
size of 500.

We show that the beam search decoder can be used to efficiently approximate a
discriminative model which alleviates exposure bias from the mismatch between
training and inference. We also avoid the label bias problem by using
unnormalized scores and performing a sequence-level normalization. Furthermore,
the use of unnormalized scores allows DBD to avoid expensive local
normalizations over large vocabularies.

Since DBD jointly trains the scoring function and the transition models, it
does away with the need for decoder hyper-parameter tuning on a held-out
validation set. We also observe on the WSJ test set that DBD can achieve better
WERs at a substantially smaller beam size (500 vs 8000) than a well tuned ASG
baseline.

On the WSJ dataset, DBD allows us to train much simpler and smaller acoustic
models with better error rates. One reason for this is that DBD can limit the
competing outputs to only sequences consisting of valid words in a given
lexicon. This frees the acoustic model from needing to assign lower probability
to invalid sequences. Including an explicit language model further decreases
the burden on the acoustic model since it does not need to learn an implicit
language model. We show that models with fewer parameters and half the temporal
receptive field can achieve equally good error rates when using DBD.

\bibliographystyle{unsrtnat}
\bibliography{refs}

\end{document}